\documentclass[letterpaper, 10 pt, journal, twoside]{ieeetran}
\usepackage{amsmath,amsfonts}
\usepackage{array}
\usepackage[caption=false,font=normalsize,labelfont=sf,textfont=sf]{subfig}
\usepackage{textcomp}
\usepackage{stfloats}
\usepackage{url}
\usepackage{verbatim}
\usepackage{graphicx}

\usepackage{hyperref}
\usepackage{graphics} 
\usepackage{times}
\usepackage{booktabs}      
\usepackage{nicefrac} 
\usepackage{microtype}
\usepackage{algorithm,multirow,xcolor}
\usepackage{algorithmicx}
\usepackage{algpseudocode}
\usepackage{mathtools}
\usepackage{cases}
\usepackage{color}
\usepackage{float}
\usepackage{bm}
\usepackage[sort]{cite}
\usepackage{tabularx}
\usepackage{threeparttable}
\usepackage{wrapfig}
\usepackage{amssymb}
\usepackage{soul}
\usepackage{epstopdf}
\usepackage{balance}
\usepackage{caption}
\usepackage{diagbox}
\def\BibTeX{{\rm B\kern-.05em{\sc i\kern-.025em b}\kern-.08em
    T\kern-.1667em\lower.7ex\hbox{E}\kern-.125emX}}
\usepackage{balance}

\title{Multi-Task Multi-Agent Reinforcement Learning via Skill Graphs}

\author{Guobin Zhu$^{1,2}$, Rui Zhou$^{1}$, Wenkang Ji$^{2}$, Hongyin Zhang$^{3}$, Donglin Wang$^{3}$ and Shiyu Zhao$^{2}$%
\thanks{Manuscript received: March, 22, 2025; Revised xxx, xxx; Accepted xxx, xxx. This paper was recommended for publication by xxx upon evaluation of the Associate Editor and Reviewers' comments.
This work was supported by the STI 2030-Major Projects (No.2022ZD0208804), National Natural Science Foundation of China (No.62473017, No.62473320). (Corresponding author: Shiyu Zhao.)} 
\thanks{$^{1}$School of Automation Science and Electrical Engineering, Beihang University, Beijing, China.}%
\thanks{$^{2}$WINDY Lab, Department of Artificial Intelligence, Westlake University, Hangzhou, China.}%
\thanks{$^{3}$MiLAB Lab, Department of Artificial Intelligence, Westlake University, Hangzhou, China.}%
\thanks{E-mail: \{zhugb, zhr\}@buaa.edu.cn, \{jiwenkang, zhanghongyin, wangdonglin, zhaoshiyu\}@westlake. edu.cn.}
\thanks{Digital Object Identifier (DOI): see top of this page.}
}

\begin{document}
\bstctlcite{IEEEexample:BSTcontrol}
\maketitle

\begin{abstract}
Multi-task multi-agent reinforcement learning (M T-MARL) has recently gained attention for its potential to enhance MARL's adaptability across multiple tasks. However, it is challenging for existing multi-task learning methods to handle complex problems, as they are unable to handle unrelated tasks and possess limited knowledge transfer capabilities. In this paper, we propose a hierarchical approach that efficiently addresses these challenges. The high-level module utilizes a skill graph, while the low-level module employs a standard MARL algorithm. Our approach offers two contributions. First, we consider the MT-MARL problem in the context of unrelated tasks, expanding the scope of MTRL. Second, the skill graph is used as the upper layer of the standard hierarchical approach, with training independent of the lower layer, effectively handling unrelated tasks and enhancing knowledge transfer capabilities. Extensive experiments are conducted to validate these advantages and demonstrate that the proposed method outperforms the latest hierarchical MAPPO algorithms. Videos and code are available at \url{https://github.com/WindyLab/MT-MARL-SG}
\end{abstract}


\section{Introduction}
\IEEEPARstart{M}{ulti-agent} reinforcement learning (MARL) is promising for decision-making in multi-robot systems and has achieved notable success in areas like games~\cite{vinyals2019grandmaster,pang2019reinforcement} and robotic control~\cite{fan2020distributed,de2021decentralized,johnson2022multi}. However, its performance tends to be task-specific, requiring retraining for new tasks due to weak generalization \cite{zhang2022discovering}. Therefore, it is important to study how MARL can adapt flexibly to multiple tasks.

In recent years, numerous studies on multi-task reinforcement learning (MTRL) have emerged~\cite{zhang2021survey,omidshafiei2017deep,hessel2019multi}. These works leverage inter-task correlations to enhance performance across tasks. Existing approaches can be broadly categorized into two types. The first focuses on learning multiple tasks simultaneously, where shared attributes/knowledge (\emph{e.g.} data, model parameters) among tasks are utilized to accelerate learning~\cite{liu2021conflict}. The second emphasizes sequential task learning, where previously acquired knowledge is retained and reused to facilitate learning when encountering related new tasks~\cite{tessler2017deep,liu2019lifelong}. Both approaches inherently involve knowledge transfer. A common strategy is to distill knowledge from multiple expert policies into a single network through supervised learning, enabling the integrated network to adapt to diverse tasks~\cite{parisotto2015actor,rusu2015policy}.
\begin{figure}
    \centering
    \includegraphics[width=1\linewidth]{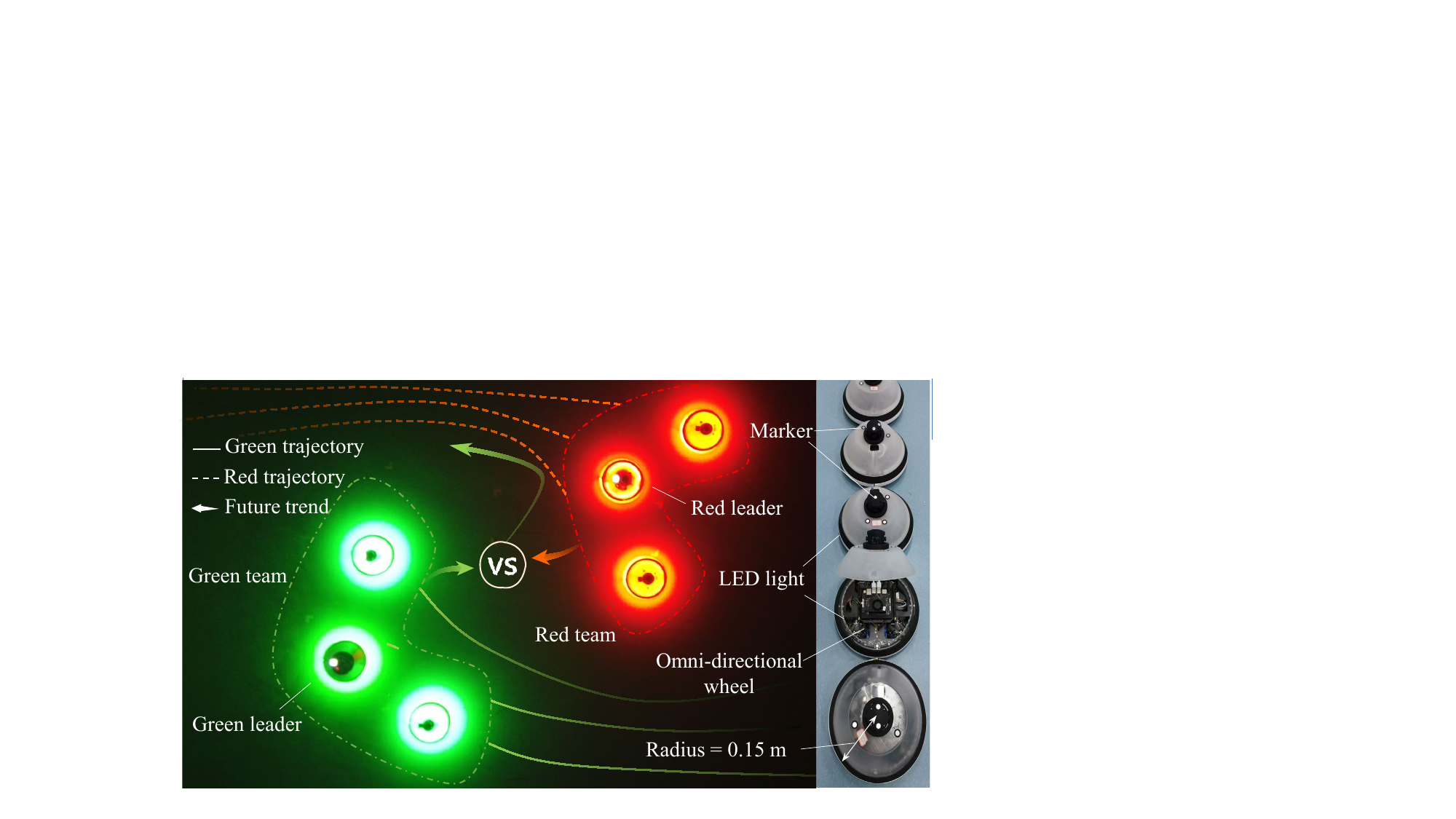}
    \caption{Real-world experiment. The red and green robot teams, moving in flocking, engage in combat. After defeating the red team, the green team continues flocking motion.}
    \label{fig:real_battle}
\end{figure}
\begin{figure*}
    \centering
    \includegraphics[width=1\linewidth]{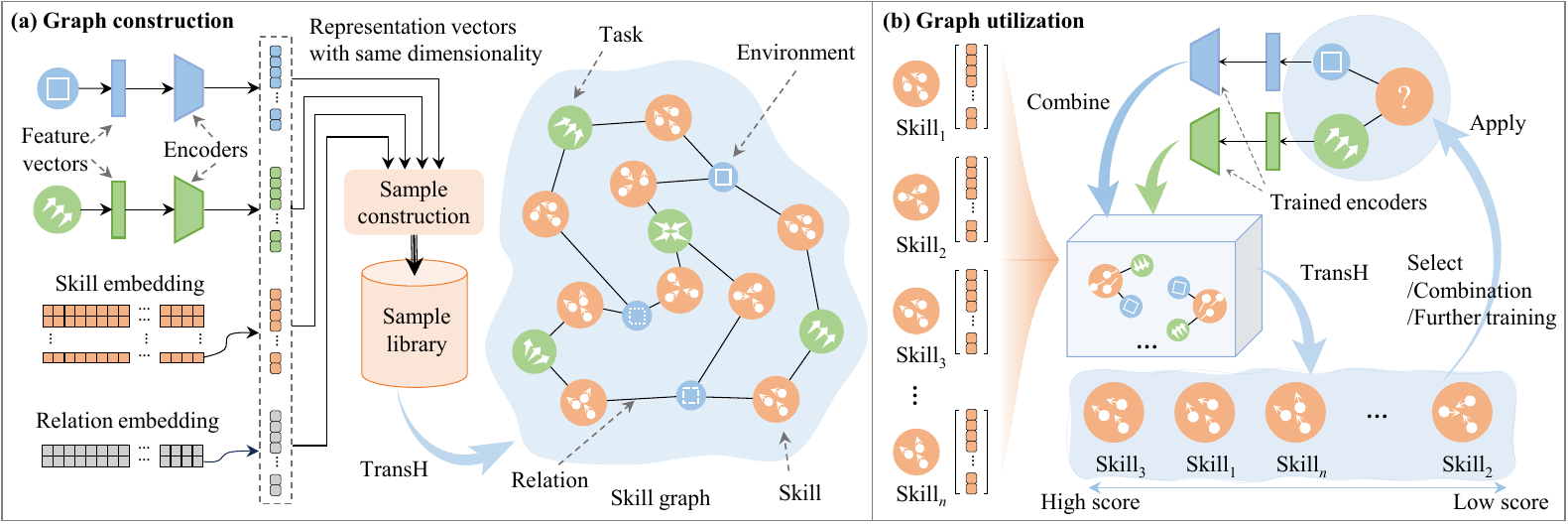}
    \caption{Method overview. (a) is the construction of the skill graph and (b) is the utilization of the skill graph. In each embedding layer and representation vector, each small square represents a numerical value.}
    \label{fig:overview}
\end{figure*}

Although MTRL has demonstrated effectiveness by leveraging inter-task knowledge~\cite{xu2020knowledge,sun2022paco}, it still has two notable limitations when applied to complex tasks. First, MTRL assumes that tasks share common attributes. If tasks are unrelated (\emph{e.g.} adversarial and cooperative), knowledge transfer becomes infeasible, rendering conventional MTRL methods ineffective. Second, current MTRL rarely accounts for multi-agent settings. Multi-agent systems involve inter-agent interactions and often operate under partial observability, making MT-MARL fundamentally different from a direct extension of single-agent MTRL.

Coincidentally, hierarchical reinforcement learning (HRL) offers a promising approach. HRL is designed to decompose complex, long-term tasks into a hierarchy of subtasks, with the high-level policy deciding which low-level skill to execute~\cite{pateria2021hierarchical,gieselmann2021planning,christen2021learning,andreas2017modular}. In multi-agent multi-task settings, HRL presents two main advantages. First, it can be naturally integrated with MTRL~\cite{tessler2017deep,andreas2017modular}. In a hierarchical framework, the high-level policy's selection of low-level skills effectively serves as a mechanism for transferring knowledge of modular skills. Since the tasks underlying these skills may be either related or unrelated, the hierarchical approach is well-suited for multi-task scenarios. Second, HRL simplifies complex multi-agent settings by breaking problems down in a layered, step-by-step manner, thereby reducing both environmental and task complexity \cite{lee2023adaptive,wang2024initial}.

However, existing HRL approaches still have two limitations. The first is that standard HRL cannot select multiple low-level strategies concurrently; in other words, HRL's capacity for knowledge transfer is limited. The second is that most HRL studies consider only several dozen agents~\cite{zhang2021multi}, and as this number increases, the curse of dimensionality becomes a significant issue. In light of these limitations, there is a need for more general MT-MARL methods. 

This paper proposes an effective hierarchical approach to address multi-task multi-agent problems. We considered the unrelated task settings, a scenario that has been rarely considered in previous work. On the high level, our approach uses a skill graph~\cite{zhang2023rsg}, inspired by the concept of knowledge graph (KG)~\cite{ji2021survey}. KG conveniently describes entities and their relationships and enables effective knowledge transfer when queried. Thus, our method consists of two steps. First, we treat environments, tasks, and skills in RL as entities and use knowledge graph embedding (KGE)\cite{wang2014knowledge} to learn representations of these entities and their relationships, thereby constructing a skill graph. Second, when queried by new environments or tasks, all skills in the graph are evaluated via a scoring mechanism, and the final skill is selected, combined, or further refined from the high-scoring ones. Such a skill graph can handle unrelated tasks and enhance effective knowledge transfer. On the low level, our approach employs the MADDPG algorithm~\cite{lowe2017multi} with a local critic instead of a centralized one. This modification better aligns with robots' local perception and supports large-scale tasks. Finally, both simulation and real-world experiments verify the effectiveness of the proposed method.

In summary, the main contributions are as follows:
1) Our approach can handle multiple unrelated tasks, including both adversarial and cooperative tasks. This unrelated case has been rarely considered in prior work since no transferable knowledge exists. The proposed approach can handle the unrelated case due to its hierarchical structure combined with a skill graph.
2) Our approach enables skill selection, combination, and further learning through a scoring mechanism, offering effective knowledge transfer compared to standard HRL. The scoring mechanism inherently offers a more robust and flexible approach to standard HRL since it always produces a list of scores, meaning that high-scoring skills that rank at the top are always obtained. 
3) We conducted experiments on a real robotic swarm platform to validate the effectiveness of the proposed method.

\section{Preliminaries}\label{sec:Preliminaries}
\subsection{Knowledge Graph}\label{subsec:Knowledge Graph}
A KG is a multi-relational graph composed of entities (nodes) and relations (different types of edges) \cite{wang2017knowledge}. Mathematically, a KG is defined as $\mathcal{G} = (\mathcal{E}, \mathcal{R}, \mathcal{T})$, where $\mathcal{E}$ is the set of all entities contained in the KG. $\mathcal{R}$ represents the set of all relations. $\mathcal{T}$ is the set of factual triples, where a standard binary fact is a triple $(h, r, b)$, $h$ and $b$ mean the head entity and the tail entity, and $r$ is the relation between the head entity and the tail entity.

Such a structure conveniently stores and links diverse knowledge with a unified specification. When encountering a new query, the KG can rapidly retrieve the target entity, while its associated entities can also be easily identified. Therefore, the KG facilitates efficient knowledge transfer \cite{xu2022novel}.

\subsection{Knowledge Graph Embedding}\label{subsec:Knowledge Graph Embedding}
Although the KG provides a convenient representation of structured knowledge, its symbolic nature poses operational challenges. KGE addresses this issue by mapping entities and relations onto a continuous vector space (\emph{i.e.}, a representation space), thereby facilitating operations while preserving the inherent structure of the KG.

The typical KGE process consists of three key steps \cite{wang2017knowledge}. First, entities and relations are represented, typically as vectors or matrices. Second, a scoring function is defined to assess the plausibility of a given fact triple, where observed facts in the KG generally receive higher scores than unobserved ones. Third, the representations of entities and relations are learned by solving an optimization problem that maximizes the plausibility of all facts. In essence, the KG is considered constructed once the representations of entities and relations have been learned.

Several KGE algorithms have been proposed, such as TransE~\cite{bordes2013translating} and TransH~\cite{wang2014knowledge}. In this study, we employ the TransH model because it supports one-to-many, many-to-one, and many-to-many relationships. For example, in a one-to-many relationship, a single head entity may correspond to multiple tail entities. In the context of RL, this corresponds to training multiple skills within a single environment.

\section{Methodology}\label{sec:Methodology}
As illustrated in Fig.~\ref{fig:overview}, the proposed method comprises two main components. The first constructs the skill graph using knowledge graph (KG) construction methods with the TransH model. The second involves utilizing the skill graph, where the TransH model scores all skills in the graph to infer the most suitable one.

\subsection{Graph Construction}\label{subsec:Construction}
As described in Section \ref{subsec:Knowledge Graph Embedding}, the construction of the skill graph consists of three steps. The first step involves representing entities and relations. In the skill graph, entities include the environment ($e$), task ($t$), and skill ($s$), while relations denote the correspondence from the environment to skill ($r_{e \rightarrow s}$) and from task to skill ($r_{t \rightarrow s}$). To map both entities and relations into a unified representation vector space, there are two sub-steps (see Fig.~\ref{fig:overview}(a)). First, we use feature vectors (different from representation vectors) to describe $e$ and $t$. Then, we use an environment encoder and a task encoder (MLP networks with Leaky-ReLU activation) to map their respective feature vectors into the representation space. Second, we employ two embedding layers (initialized orthogonally) to represent skills and relations, respectively. Each embedding layer is a matrix where rows correspond to the number of relations/skills, and columns to the dimensions of their representation vectors. In this way, entities ($e$/$t$/$s$) and relations ($r$) are represented within the same vector space.

The second step involves defining the scoring function. We adopt the TransH model as our scoring function, which is expressed as 
$\mathbb{S} = \exp (-\lambda \Vert (\mathbf{h} - \mathbf{w_r}^T\mathbf{h} \mathbf{w_r}) + \mathbf{d_r} - (\mathbf{b} - \mathbf{w_r}^T\mathbf{b} \mathbf{w_r}) \Vert),$
where $\lambda$ is a normal constant. $\mathbf{h}$ and $\mathbf{b}$ denote the head entity and tail entity representation vectors, respectively. $\mathbf{w_r}$ and $\mathbf{d_r}$ denote the relation-specific hyperplane and translation vector, respectively. The scoring function measures the plausibility of a triple $(h, r, b)$.

The third step involves learning the representation vectors. This is a supervised learning process that comprises two phases: sample construction and representation learning. In the terminology of KGE, samples are categorized as positive, negative, and soft samples \cite{wang2014knowledge}. Positive samples correspond to correct facts. During the skill collection phase (see Section \ref{sec:Skill Collection and Feature Constructon}), we know which skills are trained under specific environments and tasks, allowing us to construct factual triples $(e, r_{e \rightarrow s}, s)$ and $(t, r_{t \rightarrow s}, s)$; these triples serve as positive samples. Negative samples represent incorrect facts, which may arise from errors in the entities (\emph{e.g.} $(e, r_{e \rightarrow s}, e^{'})$, $(t, r_{t \rightarrow s}, t^{'})$) or the relations (\emph{e.g.} $(e, r_{t \rightarrow s}, s)$, $(t, r_{e \rightarrow s}, s)$). Soft samples refer to triples with varying degrees of plausibility: positive samples have a plausibility of 1, negative samples 0, and soft samples fall between 0 and 1. For example, for two similar environments $e_1$ and $e_2$, if $e_1$ corresponds to skill $s_1$, then $(e_1, r_{e \rightarrow s}, s_1)$ is a positive sample and $(e_2, r_{e \rightarrow s}, s_1)$ is a soft sample. Similarly, $(t_1, r_{t \rightarrow s}, s_1)$ is positive and $(t_2, r_{t \rightarrow s}, s_1)$ is soft. In this manner, positive, negative, and soft samples can form a sample library.

With the samples prepared, the next step is to learn representations. We use the following loss function:
\begin{equation}\label{eq:loss}
    \mathcal{L} = (\mathbb{S}_{\mathrm{posi}} - 1)^2 + (\mathbb{S}_{\mathrm{nega}} - 0)^2 + ReLU(\mathbb{S}_{\mathrm{soft}} - 1 + \delta),
\end{equation}
where, $\delta = \sum _j k_j (p_{j,1} - p_{j,2}) / \sum _j k_j$ measures the similarity between entities 1 and 2, where $k_j$ represents the weight coefficient and $p_j$ denotes the $j$-th attribute of the feature vector. We mainly adjust $k_j$ to differentiate between various types of entities. The subscript of $\mathbb{S}$ indicates the sample type. As can be seen from loss \eqref{eq:loss}, we hope that the positive sample score is 1, the negative sample score is 0, and the soft sample score is no more than $1-\delta$. The value of $\mathbb{S}$ depends on the representation vectors of entities and relations, which are obtained through the encoder and embedding layers. Therefore, minimizing $\mathcal{L}$ is to optimize the parameters of encoders and embedding layers. The result of constructing the skill graph is to obtain these networks.
\begin{algorithm}[t]
    \caption{Construction and Utilization of Skill Graph}\label{algo:sg}
    \begin{algorithmic}[1]
        \State \textbf{Construction:}
        \State Initialize encoders of $e$ and $t$, embeddings of $s$ and $r$
        \State Construct positive $(e/t, r_{e/t \rightarrow s}, s)$, negative $(e/t, r_{e/t \rightarrow s}, $ $e^{'}/t^{'})$, $(e/t, r_{t/e \rightarrow s}, s)$ and soft $(e_2/t_2, r_{e/t \rightarrow s}, s_1)$
        
        \For {iteration = $1$ to $M$}
            \State Randomly sample a mini-batch from the library
            \State Map $e$ and $t$ to representation space using encoders
            \State Update encoder and embeddings by minimizing $\mathcal{L}$
        \EndFor
        
        \State \textbf{Utilization:}
        \State Map $e_{\mathrm{new}}$, $t_{\mathrm{new}}$ to representation space using encoders
        \State Combine the representation vectors of $e_{\mathrm{new}}$, $t_{\mathrm{new}}$, $r$, $s$
        \State Compute $\mathbb{S} = \mathbb{S}_{(t, r_{t \rightarrow s}, s)} \cdot \mathbb{S}_{(e, r_{e \rightarrow s}, s)}$ for each combination
        \State Infer the most suitable skill based on scores
    \end{algorithmic}
\end{algorithm}
\begin{figure*}[t]
    \centering
    \includegraphics[width=1\linewidth]{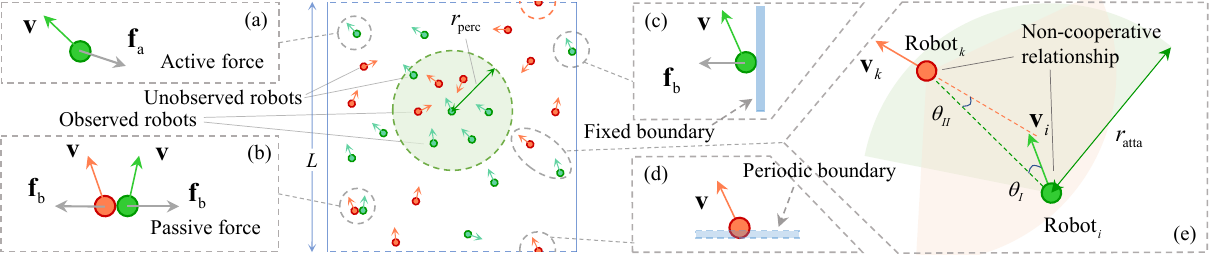}
    \caption{Motion environment illustration. (a)(b) show the active and passive force, (c)(d) shows the fixed and periodic boundaries, and (e) shows the relative motion relationship, where $\theta _I = \arccos (\mathbf{p}_{ki}\cdot \mathbf{v}_i / (\Vert \mathbf{p}_{ki} \Vert \cdot \Vert \mathbf{v}_{i} \Vert))$,$\theta _{II} = \arccos (\mathbf{p}_{ki}\cdot \mathbf{v}_k / (\Vert \mathbf{p}_{ki} \Vert \cdot \Vert \mathbf{v}_{k} \Vert))$, The light green and red sectors represent $\theta _I \leq k_{I} \pi$, $\theta _{II} \leq k_{II} \pi$ and we choose $k_{I}=k_{II}=0.4$.}
    \label{fig:environment_description}
\end{figure*}
\subsection{Graph Utilization}\label{subsec:Utilization}
With the constructed skill graph, we can identify the most suitable skill when facing a new environment $e_{\mathrm{new}}$ and task $t_{\mathrm{new}}$. The process involves three steps (see Fig.~\ref{fig:overview}(b)). First, map the feature vectors of $e_{\mathrm{new}}$ and $t_{\mathrm{new}}$ to the representation space using their trained encoders. Second, combine their representation vectors with the relation and skill representation vectors from the graph. Third, score each combination using $\mathbb{S} = \mathbb{S}_{(t, r_{t \rightarrow s}, s)} \cdot \mathbb{S}_{(e, r_{e \rightarrow s}, s)}$ and TransH model to generate a sequence of scores. The resulting scores indicate how well each skill in the graph matches the query $e_{\mathrm{new}}$ and $t_{\mathrm{new}}$.

If the score is close to 1 ($\alpha _{\mathrm{high}} < \mathbb{S} \leq 1$), it indicates that the query pair has been previously encountered; thus, the highest-scoring skill is directly applied. If the score is in the intermediate range ($\alpha _{\mathrm{low}} < \mathbb{S} \leq \alpha _{\mathrm{high}}$), it suggests a reasonable relationship between the skill and the query pair, and the robot's action is derived through $\mathbf{a} = \sum _j w_j \mathbf{a}_j$, where $j$ is the skill index for this score range, $w_j$ is the normalized score for skill $j$. Here, normalized scores mean $\sum _j w_j = 1$. If all skill scores are low ($0 \leq \mathbb{S} \leq \alpha _{\mathrm{low}}$), it indicates that the query pair falls outside the skill distribution in the graph. In this case, the highest-scoring skill is further optimized using RL and then applied to the query pair.

Throughout the construction and utilization of the skill graph, it does not require task-related dependencies and can acquire the most suitable skills through selection, combination, or further learning, thereby enhancing knowledge transfer. This approach offers an effective solution for multi-task learning. The complete process is outlined in Algorithm \ref{algo:sg}.

\section{Skill Collection and Feature Constructon}\label{sec:Skill Collection and Feature Constructon}
This section presents skill collection and feature vector construction through specific environments and task examples. It first introduces the environment, then describes adversarial and cooperative tasks—highlighting the proposed method's ability to handle unrelated tasks—and finally outlines the MARL algorithm and feature vector construction.

\subsection{Environment Description}\label{subsec:Environment Description}
The environment is shown in Fig.~\ref{fig:environment_description}, where the robot is represented as a disk. Robots of the same/different colors represent cooperative/non-cooperative relationships. Taking the green side as an example, consider $N$ homogeneous robots, defined as $\mathcal{A}=\{1,\ldots, N\}$. Each robot's state is $\mathbf{x} = [\mathbf{p}^{T}, \mathbf{v}^{T}]^{T}$, where, $\mathbf{p}$ and $\mathbf{v}$ are the position and velocity, respectively. The robot's movement is driven by the active and passive force (see Fig.~\ref{fig:environment_description}(a)(b)). The active force, $\mathbf{f}_{\mathrm{a}}$, is a self-generated force, which is the output of the actor network. The passive force, $\mathbf{f}_{\mathrm{b}}$, is an elastic force following Hooke's Law. Thus, the robot's dynamics is
$\dot{\mathbf{p}}_i = \mathbf{v}_i, \ \dot{\mathbf{v}}_i = (\mathbf{f}_{\mathrm{a}} + \mathbf{f}_{\mathrm{b}})/m_i, \ i \in \mathcal{A}$,
where $m_i$ is the mass of robot $i$, and the velocity is bounded by $v_{\mathrm{min}} \leq \Vert \mathbf{v}_i \Vert \leq v_{\mathrm{max}}$. 

The robot's perception range is a circular area with radius $r_{\mathrm{perc}}$. In the observation model, perception depends on both Euclidean and topological distances~\cite{li2023predator}, where the topological component limits the number of perceivable individuals. We combine these factors and set a threshold of $n_{\mathrm{h}} = 6$~\cite{li2023predator}. 

All robots move within a bounded region of length $L$. We consider two types of boundaries: a fixed boundary (see Fig.~\ref{fig:environment_description}(c)), where robots cannot pass through and will be repelled upon contact, and a periodic boundary (see Fig.~\ref{fig:environment_description}(d)), where robots can exit the environment from one side and reappear on the opposite side at the same velocity.

\subsection{Adversarial Task}\label{subsec:Adversarial Task}
The adversarial task involves two groups of robots, where each side aims to defeat the other. The task ends when one group is eliminated. The adversarial task setup mirrors many biological group behaviors~\cite{dorigo2021swarm}, with the attack process illustrated in Fig.~\ref{fig:environment_description}(e). From the green side's perspective, a red robot is considered attacked if the following conditions are met: 1) the green robot is behind the red robot (the light red area); 2) the green robot's velocity is approximately directed towards the red robot (the light green area); and 3) the distance between them is less than the attack radius $r_{\mathrm{atta}}$.

In our setup, each robot has a health value $hp$. When a robot is attacked by an enemy (non-cooperating robot), its $hp$ decreases by $\triangle h$, with a maximum of $n_{\mathrm{o}}$ enemies attacking simultaneously. Additionally, if a robot is not under attack, its $hp$ increases by $0.1\triangle h$ up to the initial health value $hp_{\mathrm{max}}$. A robot is considered dead when $hp \leq 0$.

1) \emph{Reward}: The reward can be expressed as $r_{\mathrm{a}} = r_{\mathrm{surv}} + r_{\mathrm{situ}}$, with default values $r_{\mathrm{surv}} = r_{\mathrm{situ}} = 0$. Here, $r_{\mathrm{surv}}$ represents the survival reward. If a robot eliminates an enemy, $r_{\mathrm{surv}} = k_{\mathrm{surv}}$; if it is eliminated, $r_{\mathrm{surv}} = -k_{\mathrm{surv}}$. $r_{\mathrm{situ}}$ represents the situational reward if a robot meets the conditions to attack an enemy, $r_{\mathrm{situ}} = k_{\mathrm{situ}}$. Both $k_{\mathrm{surv}}$ and $k_{\mathrm{situ}}$ are positive constants.

2) \emph{Action}: The action is a two-dimensional vector with components along the x and y axes. This vector indicates the active force $\mathbf{f}_{\mathrm{a}}$ exerted on the robot.

3) \emph{Observation}: Based on the setup, each robot's observation vector is designed to include its own state and the relative states of teammates and enemies. We set the maximum number of teammates and enemies to $n_{\mathrm{h}}/2$. For example, the robot $i$'s observation is $\mathbf{o}_i=[\mathbf{x}_i^T, \mathbf{x}_{ji,1}^T, \ldots, \mathbf{x}_{ji,n_{\mathrm{h}}/2}^T, \mathbf{x}_{ki,1}^T \ldots, $ $ \mathbf{x}_{ki,n_{\mathrm{h}}/2}^T]^T$, where, $\mathbf{x}_{ji} = \mathbf{x}_j - \mathbf{x}_i, j \in \mathcal{N}_i$, $\mathbf{x}_{ki} = \mathbf{x}_k - \mathbf{x}_i, k \in \mathcal{O}_i$, $\mathcal{N}_i$/$\mathcal{O}_i$ are the sets of robot $i$'s teammates/enemies. 

\subsection{Cooperative Task}\label{subsec:Cooperative Task}
The cooperative task involves a single group of robots, for which we select flocking—a behavior widely observed in nature~\cite{dorigo2021swarm, vasarhelyi2018optimized}. Flocking is typically governed by three heuristic rules: \emph{cohesion}, \emph{separation}, and \emph{alignment}. A robot is attracted to or repelled by neighbors when their distance is too large or too small, respectively, and its velocity tends to align with the average of its neighbors. Ultimately, the inter-robot distance stabilizes at a certain value.

1) \emph{Reward}: Similar to the three flocking rules, we designed three reward criteria to achieve the same effect. $r_{\mathrm{attr}} = -k_{\mathrm{attr}}(d_{ij} - d_{\mathrm{ref}}) \ \text{if} \ d_{ij} > d_{\mathrm{ref}}$, and 0 otherwise. $r_{\mathrm{repl}} =-k_{\mathrm{repl}}(d_{\mathrm{ref}} - d_{ij}) \ \text{if} \ d_{ij} < d_{\mathrm{ref}}$, and 0 otherwise. $r_{\mathrm{alig}} = -k_{\mathrm{alig}}\Vert \frac{1}{N_i} \sum _{j \in \mathcal{N}_i} (\frac{\mathbf{v}_j}{\Vert \mathbf{v}_j \Vert} - \frac{\mathbf{v}_i}{\Vert \mathbf{v}_i \Vert}) \Vert$,
where $k_{\mathrm{attr}}$, $k_{\mathrm{repl}}$, $k_{\mathrm{alig}}$ are positive constants, $d_{ij} = \Vert \mathbf{p}_i - \mathbf{p}_j \Vert$ is the distance between $i$ and $j$, $d_{\mathrm{ref}}$ is the final stable value, and $N_i$ is the number of robot $i$'s neighbors. Therefore, the total reward is expressed as
$r_{\mathrm{f}} = r_{\mathrm{attr}} + r_{\mathrm{repl}} + r_{\mathrm{alig}}$.

2) \emph{Action}: Flocking involves only the robot's movement; therefore, the action is the same as in the adversarial task.

3) \emph{Observation}: Each robot's observation vector is designed to include its own state and the relative states of its neighbors. For example, the observation of robot $i$ is denoted as $\mathbf{o}_i=[\mathbf{x}_i^T, \mathbf{x}_{ji,1}^T, \ldots, \mathbf{x}_{ji,n_{\mathrm{h}}}^T]^T$.

\subsection{MARL Algorithm}\label{subsec:MARL Algorithm}
This paper uses a local-critic version of MADDPG (local-MADDPG) to train the low-level skills. MADDPG employs an actor-critic architecture~\cite{lowe2017multi}. In this paper, both the actor and critic are MLPs, with Leaky-ReLU in hidden layers, Tanh for the actor's output, and no activation for the critic's output. As the robots are homogeneous, a single actor and critic can be shared among cooperative robots. Specifically, the adversarial task involves two critics and actors, while flocking involves one of each.

The centralized critic in~\cite{lowe2017multi} is $Q(\mathbf{x}_1,\ldots, \mathbf{x}_N, \mathbf{a}_1,\ldots,$ $ \mathbf{a}_N)$, where, $\mathbf{a}_i = \mu(\mathbf{o}_i), i \in \mathcal{A}$ and $\mu$ represents the actor. We modified it to $Q(\mathbf{o}_i, \mathbf{a}_i)$; thus, the critic only inputs the robot $i$'s observation and action. This mimics the local perception in biological swarms, while the local critic enables efficient handling of large-scale tasks. For more details on MADDPG, please refer to~\cite{lowe2017multi}. 

\subsection{Feature Construction}\label{subsec:Feature Construction}
In this paper, feature vectors serve as numerical representations of entities (\emph{i.e.}, environments or tasks), capturing their attributes. An environment can be represented as $e=(y, L)$, where $y$ indicates the boundary type, with $y = 0$ representing a periodic boundary and $y = 1$ representing a fixed boundary. An adversarial task can be represented as $t=(v_{\mathrm{max}}, v_{\mathrm{min}}, \triangle h, n_{\mathrm{o}}, r_{\mathrm{atta}})$, and a flocking task can be represented as $t=(v_{\mathrm{max}}, v_{\mathrm{min}}, d_{\mathrm{ref}}, r_{\mathrm{perc}})$. Once these feature vectors are constructed, they are encoded as representation vectors through environment or task encoders. By combining the skill and relation embedding layers, the skill graph can then be constructed.

\begin{figure*}[!t]
    \centering
    \includegraphics[width=1\linewidth]{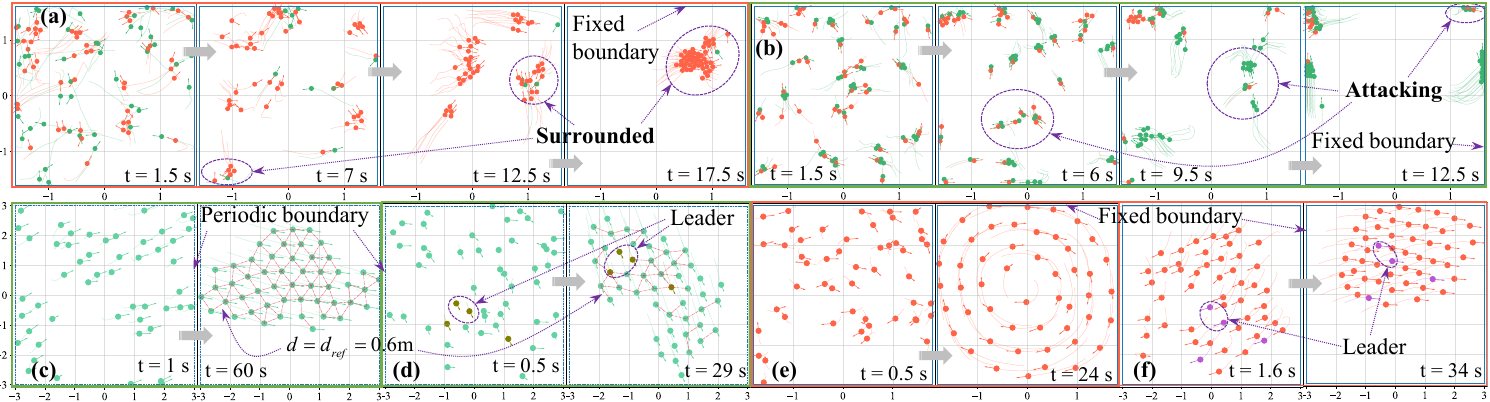}
    \caption{Low-level skill collection. (a) and (b) illustrate an adversarial task where the green team uses local-MADDPG, and the red team employs an auxiliary strategy with $k_{\mathrm{p}} = 1$ and $k_{\mathrm{v}} = 2$. In (a), local-MADDPG is untrained, leading to the green team's defeat. In (b), after training, the green team prevails over the red team. (c)(d), (e)(f) illustrate the flocking behavior with local-MADDPG strategy under periodic and fixed boundaries, respectively.}
    \label{fig:flocking_and_adversarial}
\end{figure*}

\section{Simulation Experiments}\label{sec:Simulation Analysis}
This section delineates three components: the collection of low-level skills, the application of the skill graph, and a direct comparative analysis between the skill graph and HRL. The analysis aims to elucidate the advantages of integrating hierarchical structures with the skill graph, thereby contrasting this approach with the standard HRL approach. Moreover, HRL adopts the latest MAPPO algorithm \cite{yu2022surprising}. We carefully considered alternative baselines, but most existing methods are not directly applicable to the unrelated tasks in our study. Thus, they do not offer a fair basis for comparison.

\subsection{Low-level Skill Collection}\label{subsec:Low-level Skill Collection}
1) \emph{Experimental setup}: 
The number of robots for both teams in the adversarial task and for the flocking task is set to $N=50$. In the adversarial task, the red team adopts an auxiliary strategy $\mathbf{f}_{\mathrm{a}} = k_{\mathrm{p}}(\mathbf{p}_{\mathrm{c}} - \mathbf{p}) + k_{\mathrm{v}}(\mathbf{v}_{\mathrm{c}} - \mathbf{v})$, where $\mathbf{p}_{\mathrm{c}}$ and $\mathbf{v}_{\mathrm{c}}$ are the position and velocity of the nearest green robot. The green team adopts the local-MADDPG strategy. In the flocking task, to prevent swarm splitting due to local perception, several robots are designated as leaders with fixed movements (see Fig.~\ref{fig:flocking_and_adversarial}(d)(f)). These leaders are otherwise identical to regular agents. The task parameters are set as follows: $r_{\mathrm{perc}} = 3 \ \mathrm{m}$, $m_i = 1 \ \mathrm{kg}$, $v_{\mathrm{max}} = 1 \ \mathrm{m/s}$, $v_{\mathrm{min}} = 0$, $L = 4 \sim 6 \ \mathrm{m}$, $\triangle h = 1$, $hp_{\mathrm{max}} = 80$, $n_{\mathrm{o}} = 3$, $r_{\mathrm{atta}} = 0.3 \ \mathrm{m}$, $d_{\mathrm{ref}} = 0.4 \sim 0.7 \ \mathrm{m}$, $k_{\mathrm{situ}} = 1$, $k_{\mathrm{surv}} = 5$, $k_{\mathrm{attr}} = 1 \sim 2.5$, $k_{\mathrm{repl}} = 15$, $k_{\mathrm{alig}} = 2$. Hyperparameters for local-MADDPG are listed in Table~\ref{tab:hyper_parameter}.

2) \emph{Results analysis}: 
The experimental results are shown in Fig.~\ref{fig:flocking_and_adversarial}. In the adversarial scenario, the green team's performance depends on whether its local-MADDPG policy is trained. When untrained, the red team—guided by an auxiliary strategy—successfully maneuvers behind and eliminates it (see Fig.~\ref{fig:flocking_and_adversarial}(a)). Conversely, a trained green team can outmaneuver and defeat the red team. (see Fig.~\ref{fig:flocking_and_adversarial}(b)). 
In the flocking task (see Fig.~\ref{fig:flocking_and_adversarial}(c)-(f)), the robot team maintains nearly equal inter-agent distances once a stable formation is reached. Notably, under periodic boundary conditions, Reynolds' three rules are well preserved. These results demonstrate that the local-MADDPG algorithm successfully collects low-level skills.

\subsection{Skill Graph Application}\label{subsec:Skill Graph Application}
1) \emph{Experimental setup}: 
Suppose we have 8 flocking subtasks with feature vectors as follows: floc\_1: (1,0,0.4,2), floc\_2: (1,0,0.8,2), floc\_3: (1,0,0.4,3), floc\_4: (1,0,0.8,3), floc\_5: (1,0,0.4,4), floc\_6: (1,0,0.8,4), floc\_7: (1,0,0.4,5), floc\_8: (1,0,0.8,5). Additionally, there are 8 adversarial subtasks with feature vectors: adve\_1: (1,0,1,2,0.3), adve\_2: (1,0,1,3,0.3), adve\_3: (1,0,1,4,0.3), adve\_4: (1,0,1,5,0.3), adve\_5: (1,0,1,2, 0.4), adve\_6: (1,0,1,3,0.4), adve\_7: (1,0,1,4,0.4), adve\_8: (1,0, 1,5,0.4). Moreover, there are 2 environments: fixed: (1,6) and periodic: (0,6). This yields a total of 32 skills (\emph{i.e.}, (8+8)$\times$ 2 = 32). Using these tasks, environments, and skills, we can construct samples and subsequently construct a skill graph. The training parameters for the skill graph are set as follows: representation vector size = 96, hid size = 256, hid layers = 3, $M$ = 500, batch size = 256, $k_{\mathrm{1,t}} = 0$, $k_{\mathrm{2,t}} = 0$, $k_{\mathrm{3,t}} = 0$, $k_{\mathrm{4,t}} = 3$, $k_{\mathrm{5,t}} = 1$, $k_{\mathrm{1,e}} = 0.95$, $k_{\mathrm{2,e}} = 0.05$, $\lambda = 3$, $\alpha_{\mathrm{high}} = 0.95$, $\alpha_{\mathrm{low}} = 0.85$. 
\begin{table}[t]
    \centering
    \caption{Hyperparameters for local-MADDPG and MAPPO}
    \begin{threeparttable}
    \setlength{\tabcolsep}{0.1em}{
    \renewcommand\arraystretch{1.2}
    \begin{tabular}{llll|llll}
        \toprule
         \multicolumn{4}{c}{local-MADDPG} & \multicolumn{4}{c}{MAPPO\cite{yu2022surprising}} \\
        \cmidrule(lr){1-4} \cmidrule(lr){5-8}
        & Adve/Floc & & Adve/Floc & & Adve/Floc & & Adve/Floc \\
        \midrule
        eps & 1200/1800 & critic lr & 1e-3/1e-3 & - & 2400/2000 & - & 5e-4/1e-3 \\
        eps len & 200/200 & actor lr & 1e-4/1e-4 & - & 350/200 & - & 5e-4/1e-3 \\
        buf size & 5e5/5e5 & gamma & 0.99/0.99 & - & 1e5/1e5 & - & 0.99/0.99 \\
        batch & 512/512 & explr rate & 0.1/0.1 & - & 2048/2048 & ppo epoch & 20/20 \\
        hid size & 128/64 & noise & 0.8/0.8 & - & 256/128 & clip para & 0.1/0.2 \\
        hid layers & 2/3 & soft-upd & 0.01 & - & 3/3 & gae gamm & 0.95/0.95 \\
        \bottomrule
    \end{tabular}
    }
    \end{threeparttable}
    \label{tab:hyper_parameter}
\end{table}

\begin{figure*}[!t]
    \centering
    \includegraphics[width=1\linewidth]{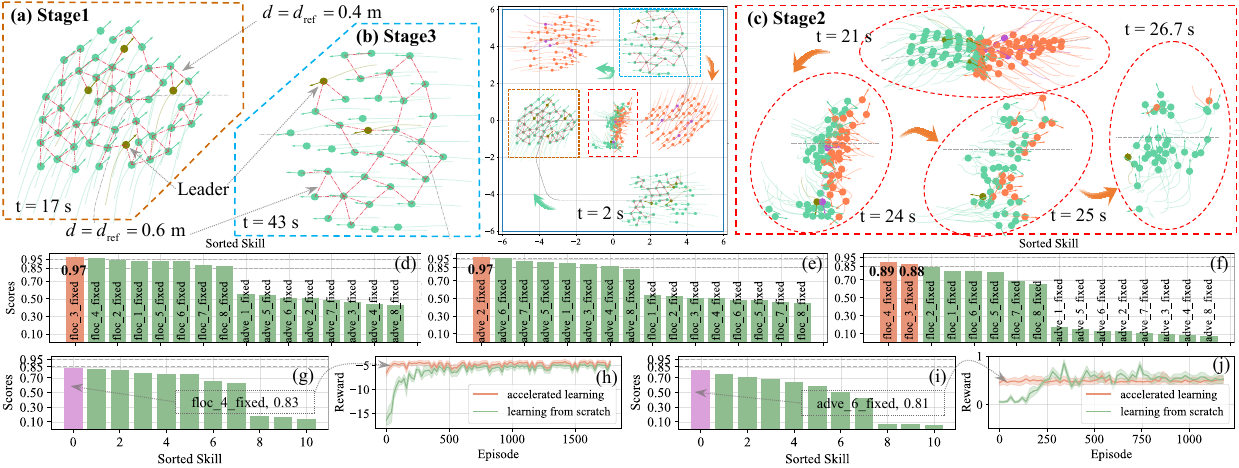}
    \caption{Skill graph application. In (a)(b), the red lines on the green side indicate that the inter-robot distance is approximately $d_{\mathrm{ref}}$. In (c), the green side uses local-MADDPG strategy, while the red side employs an auxiliary strategy with $k_{\mathrm{p}} = 1$ and $k_{\mathrm{v}} = 2$. (d)(e)(f) show the top 16 skill scores for the three stages, with red highlighting the skills that can be directly reused. (g) displays the top 11 skills for the query pair (1,6)+(1,0,1,3), while (h) illustrates the reward curves for further training of the highest-scoring skills versus training from scratch. Similarly, (i)(j) correspond to (1,6)+(1,0,1,3,0.8).}
    \label{fig:skill_graph}
\end{figure*}

To illustrate the application of the skill graph, we designed a scenario as shown in Fig.~\ref{fig:skill_graph}(a)-(c) where two groups of 50 robots each move along predetermined paths. When the groups do not encounter each other, they perform a flocking task with $d_{\mathrm{ref}} = 0.4 \ \mathrm{m}$ (stage 1). Upon encountering—defined as a robot from one group coming within $r_{\mathrm{perc}}$ of the opposing group's center—they engage in an adversarial task (stage 2). After one group is eliminated, the remaining group resumes the flocking task (stage 3) with a different $d_{\mathrm{ref}} = 0.6 \ \mathrm{m}$. In such a scenario, the skill graph needs to infer (decide) the most suitable skill for the robot team in each stage.

2) \emph{Results analysis}:
The skill graph's application is illustrated in Fig.~\ref{fig:skill_graph}. In stage 1 (see Fig.~\ref{fig:skill_graph}(a)), both robot teams perform the flocking task with $d_{\mathrm{ref}}=0.4 \ \mathrm{m}$. The input environment + task query is (1,6) + (1,0,0.4,3), and the output (see Fig.~\ref{fig:skill_graph}(d)) shows that the skill ``floc\_3\_fixed" achieves the highest score ($0.97$), exceeding the threshold $\alpha_{\mathrm{high}} = 0.95$, and is therefore selected for stage 1. Similarly, the query is (1,6) + (1,0,1,3,0.3) in stage 2, and the highest-scoring skill ``adve\_2\_fixed" is chosen (see Fig.~\ref{fig:skill_graph}(e)). In stage 3, the query is (1,6) + (1,0,0.6,3). Although this query does not match any fundamental skill in the graph, the graph provides the two most relevant skills, ``floc\_4\_fixed" and ``floc\_3\_fixed" (see Fig.~\ref{fig:skill_graph}(f)). Then, the robot's action is a weighted combination of these two skills. This process clearly demonstrates the skill graph's selection and combination capabilities. 

Fig.~\ref{fig:skill_graph}(g)-(j) further demonstrates how the skill graph handles unfamiliar queries—not entirely unseen, but rather cases that may resemble previously encountered ones. For instance, when the query input is (1,6) + (1,0,1,3), the output's highest score is below $\alpha _{\mathrm{low}}$ (see Fig.~\ref{fig:skill_graph}(g)). In this case, the highest-scoring skill, ``floc\_4\_fixed", is further trained and used for the query pair. This approach results in a skill that requires less training time and higher sampling efficiency compared to training from scratch (see Fig.~\ref{fig:skill_graph}(h)), enabling robots to quickly adapt to unfamiliar environments and tasks. In summary, the skill graph demonstrates excellent knowledge transfer capabilities.

\subsection{Comparative Experiments}\label{subsec:Comparative Experiments}
1) \emph{Experimental setup}: 
The setup of the skill graph is described in Section \ref{subsec:Skill Graph Application}; here, we focus on the HRL setup. It consists of two layers, both using MAPPO. The low-level MAPPO shares the same observation, action, and reward structure as in Section \ref{sec:Skill Collection and Feature Constructon}, with hyperparameters listed in Table \ref{tab:hyper_parameter}. The high-level MAPPO is set as follows. \emph{Observation}: For the green side, the observation vector is designed as $\mathbf{o}_{\mathrm{g}} = [\mathbf{x}_{\mathrm{rg}}^T, n_{\mathrm{g}}, n_{\mathrm{r}}]^T$, where, $\mathbf{x}_{\mathrm{rg}} = \mathbf{x}_{\mathrm{r}} - \mathbf{x}_{\mathrm{g}}$ represents the relative position between team centers, and $n_{\mathrm{g}}$ and $n_{\mathrm{r}}$ denote the number of robots on the green and red teams, respectively. \emph{Action}: The action space is discrete, matching the number of available low-level skills; the actor outputs the index of the selected skill. \emph{Reward}: A -1 reward is given if the selected skill does not match the expected one. All other high-level MAPPO hyperparameters follow those used in the flocking task (Table~\ref{tab:hyper_parameter}), except for the number of episodes (set to 200) and the episode length (set to 300).

To implement the comparison, we define two metrics. 
The first is the \emph{decision success rate}, denoted as $\rho _{\mathrm{succ}} = n_{\mathrm{succ}}/$ $n_{\mathrm{total}}$, where, $n_{\mathrm{succ}}$ and $ n_{\mathrm{total}}$ represent the numbers of successful and total decisions, respectively. we aim for the method to achieve as high a success rate as possible. 
The second is \emph{generalization} which is a qualitative metric. During testing, we will vary certain parameters and observe changes in $\rho _{\mathrm{succ}}$. Ideally, an effective method should exhibit stable performance despite these parameter variations.

It is crucial to clarify that our comparison involves two schemes: Scheme 1 (a high-level skill graph + a low-level local-MADDPG) and Scheme 2 (a high-level MAPPO + a low-level MAPPO). Despite differences in low-level algorithms, this comparison validly demonstrates the advantages of the skill graph for two reasons: First, the defined metrics do not depend on the low-level algorithm and exclusively evaluate high-level performance. Second, both frameworks are modular; although replacing Scheme 2's low-level component with the local-MADDPG is feasible, such a hybrid configuration would be less natural.
\begin{table}
    \centering
    \caption{$\rho _{\mathrm{succ}}$ of high-level MAPPO}
    \begin{threeparttable}
    \setlength{\tabcolsep}{0.15em}{
    \renewcommand\arraystretch{1.2}
    \begin{tabular}{l*{5}{l}}
        \toprule
        \diagbox{dim}{case} & base & init num*1.1 & init num*0.9 & leader v*1.4 & leader v*0.6 \\
        \midrule 
        dim=2  & 95.2(6.8)\% & 93.2(8.2)\% & 93.9(6.9)\% & 93.8(7.9)\% & 95.1(6.3)\% \\
        dim=4  & 94.6(7.4)\% & 93.1(8.2)\% & 90.9(9.1)\% & 93.2(8.7)\% & 94.3(7.7)\% \\
        dim=6  & 94.4(7.2)\% & 93.0(8.3)\% & 89.7(8.0)\% & 92.9(8.5)\% & 94.1(7.1)\% \\
        dim=8  & 93.7(7.7)\% & 91.4(8.8)\% & 88.3(8.8)\% & 92.6(8.7)\% & 93.1(8.4)\% \\
        dim=10 & 93.2(7.5)\% & 90.1(7.8)\% & 88.0(9.2)\% & 92.1(8.6)\% & 93.0(7.7)\% \\
        dim=12 & 92.5(7.6)\% & 90.7(7.9)\% & 86.8(9.5)\% & 90.7(8.8)\% & 92.1(8.0)\% \\
        \bottomrule
    \end{tabular}
    }
    \end{threeparttable}
    \label{tab:high_ppo_performance}
\end{table}

2) \emph{Results analysis}: 
Table \ref{tab:high_ppo_performance} summarizes detailed $\rho _{\mathrm{succ}}$ metric, where "dim" indicates the number of low-level skills, and "base" represents the case with identical training and testing parameters. Cols 2-5 (different training and testing parameters) represent variations in the initial number (``init num'') and leader speed (``leader v''), which aim to assess the generalization ability of the MAPPO strategy. The table demonstrates two key findings: First, $\rho _{\mathrm{succ}}$ decreases as the number of available skills increases. Second, $\rho _{\mathrm{succ}}$ is sensitive to parameter changes, indicating poor generalization.

Many parameters can highlight hierarchical MAPPO's weak generalization, but we report only two due to space limits. Its underperformance stems from two reasons. First, the high-level strategy depends on low-level skills during training, creating a tight coupling where many variations in low-level parameters can impact high-level decision-making. Second, the high-level strategy only selects or rejects low-level skills, which is too rigid and overlooks the inherent rationality of those skills. In contrast, the skill graph has three advantages. First, its construction is independent of low-level skills, ensuring that changes in low-level parameters do not affect it, leading to good generalization. Second, the skill graph selects skills based on their scores without requiring a perfect score, (\emph{i.e.} 1), resulting in $\rho_{\mathrm{succ}}=100\%$. Third, the skill graph can combine or further optimize high-scoring skills, a capability that hierarchical MAPPO lacks, \emph{i.e.} the skill graph exhibits better knowledge transfer than hierarchical MAPPO.
\begin{figure}[t]
    \centering
    \includegraphics[width=1\linewidth]{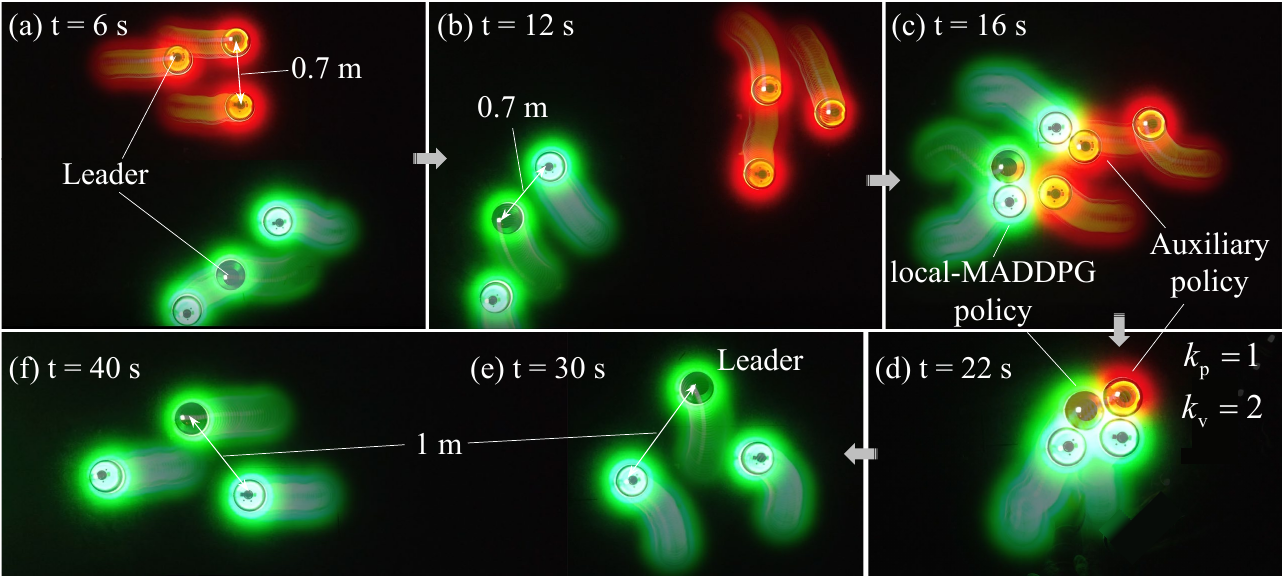}
    \caption{Snapshots of the real-world experiment. (a)(b), (c)(d), and (e)(f) correspond to stage 1, stage 2, and stage 3, respectively.}
    \label{fig:real_snapshot}
\end{figure}

\section{Real-world Experiments}\label{sec:Real-world Experiments}
The real experimental setup is shown in Fig.~\ref{fig:real_battle} and Fig.~\ref{fig:real_snapshot}, where six omnibot robots~\cite{ma2024omnibot} are divided into red and green teams, moving within a $5\times 5 \ \mathrm{m}$ area. The experiment follows the procedure illustrated in Fig.~\ref{fig:skill_graph}(a)-(c). The robot structure is depicted in Fig.~\ref{fig:real_battle}, and its state is tracked using a motion capture system. The robot skill library is mostly consistent with that mentioned in Section \ref{sec:Simulation Analysis}-B, with three differences. First, the flocking's $d_{\mathrm{ref}}$ is adjusted from 0.4/0.8 m to 0.7/1.3 m. Second, the boundary length is reduced from 6 m to 5 m. Third, training episodes for flocking are increased from 1200 to 3000. We set $d_{\mathrm{ref}}$ to $0.7 \ \mathrm{m}$ for stage 1 and $1 \ \mathrm{m}$ for stage 3. All other parameters match those in the Simulation.

In stages 1, 2, and 3, the query pairs are (1,5) + (1,0,0.7,3), (1,5) + (1,0,1,3,0.3), and (1,5) + (1,0,1,3), respectively, and the skill graph outputs skills ``floc\_3\_fixed", ``adve\_2\_fixed", and ``floc\_4\_fixed" and ``floc\_3\_fixed" for robot team. As shown in Fig.~\ref{fig:real_snapshot}, both the red and green teams first form flocking with $d_{\mathrm{ref}} = 0.7 \ \mathrm{m}$, then enter the adversarial phase, where the green team defeats the red team. Finally, the green team transitions to a flocking with $d_{\mathrm{ref}} = 1 \ \mathrm{m}$. The entire process demonstrates the practical feasibility of the method presented in this paper.

\section{Conclusion}\label{sec:Conclusion}
This paper proposed an effective hierarchical approach for MT-MARL, with a skill graph at the upper layer. The skill graph demonstrated three advantages: it did not require related attributes among low-level tasks, offered superior knowledge transfer compared to standard hierarchical methods, and was trained independently of low-level skills with good generalization. Simulations and real-world experiments were conducted to validate the advantages of the proposed approach.

It is worth noting that the fact that the high-level skill graph does not require lower-level tasks to be related does not imply that it can transfer knowledge in entirely unseen scenarios—these are two separate issues. When the latter occurs, the scores of all skills in the library tend to be low due to the absence of transferable knowledge, and new skills must be learned from scratch. This also represents a potential direction for future research.

\bibliographystyle{IEEEtran}
\bibliography{IEEEabrv,main}

\end{document}